\documentclass{article} 

\usepackage{iclr2027_conference,times}

\iclrfinalcopy           

\usepackage{amsmath,amsfonts,bm}

\def\eqref#1{equation~\ref{#1}}

\def\1{\bm{1}}

\DeclareMathAlphabet{\mathsfit}{\encodingdefault}{\sfdefault}{m}{sl}
\SetMathAlphabet{\mathsfit}{bold}{\encodingdefault}{\sfdefault}{bx}{n}

\usepackage{hyperref}
\usepackage{url}

\usepackage{booktabs}       
\usepackage{colortbl}       
\usepackage{amsfonts}       
\usepackage{nicefrac}       
\usepackage{microtype}      
\usepackage{xcolor}         

\usepackage{amsmath}
\usepackage{amssymb}
\usepackage{enumitem}
\usepackage{graphicx}
\usepackage{multirow}

\usepackage[ruled,noend,linesnumbered]{algorithm2e}

\usepackage{bbm}

\usepackage{caption}
\usepackage{listings}
\usepackage{xcolor}
\usepackage{upquote}

\usepackage{subcaption}

\usepackage{cleveref}

\newcommand{\method}{\textsc{Self-Step Rejection}}
\newcommand{\methodabbr}{\textsc{SSR}}

\DeclareCaptionFont{white}{\color{white}}
\DeclareCaptionFormat{listing}{%
  \parbox{\linewidth}{\colorbox{gray}{\parbox{\linewidth}{#1#2#3}}\vskip-4pt}}
\title{Can Language Models Learn to Reject Their Own Bad Reasoning Steps?}

\author{
Siheng Xiong$^{1}$, Xiaoze Liu$^{2}$, Yiqiao Jin$^{1}$, Xiaoqian Wang$^{2}$, Jing Gao$^{2}$\\
$^{1}$Georgia Institute of Technology \qquad $^{2}$Purdue University\\
\texttt{\{sxiong45,yjin328\}@gatech.edu,}\\
\texttt{\{xiaoze,joywang,jinggao\}@purdue.edu}
}

\begin{document}

\maketitle
\lhead{Preprint}

\begin{abstract}
Verifier-guided decoding can prevent harmful reasoning steps from contaminating
subsequent generation, but typically relies on an external learned verifier.
Can a language model instead learn to reject its \emph{own} bad reasoning steps?
This requires distinguishing candidate steps sampled from the same problem and
reasoning prefix.
We define a prefix's \emph{recoverability} as the probability that the frozen
generator can complete it correctly.
Diagnostic measurements show that most adjacent changes in recoverability are
difficult to resolve with practical Monte Carlo budgets, even when cumulative
degradation is detectable.
Meanwhile, candidates sampled from a shared prefix exhibit a sparse
low-recoverability tail, suggesting an opportunity for selective rejection.
We introduce \method{} (\methodabbr), which trains a lightweight LoRA
acceptance gate on the generator's own backbone while keeping the base model
frozen.
\methodabbr{} uses confidence-qualified \emph{first-passage supervision}:
steps before the first resolved crossing of a root-relative recoverability
barrier are labeled as accepted, the crossing step is labeled as rejected, and
unresolved steps and the subsequent suffix are excluded from supervision.
Training combines pointwise classification with same-prefix pairwise learning,
followed by group-relative policy refinement using final-answer correctness.
At inference, the gate accepts candidates or resamples from the unchanged
prefix, subject to a per-position attempt cap and an episode-level rejection
budget, without an external learned verifier.
Across three reasoning models and five mathematical reasoning benchmarks,
\methodabbr{} improves macro-average accuracy over single-pass decoding by
5.4--10.1 percentage points using 1.21--1.40$\times$ as many generated tokens,
achieving the highest macro-average accuracy among the evaluated step-level
methods for each generator, while full-solution scaling methods require
4.47--8.27$\times$ the single-pass token cost for comparable performance.
The code and model checkpoints are available at \url{https://github.com/xiongsiheng/SSR}.
\end{abstract}

\begin{figure*}[h]
    \centering
    \includegraphics[width=\textwidth]{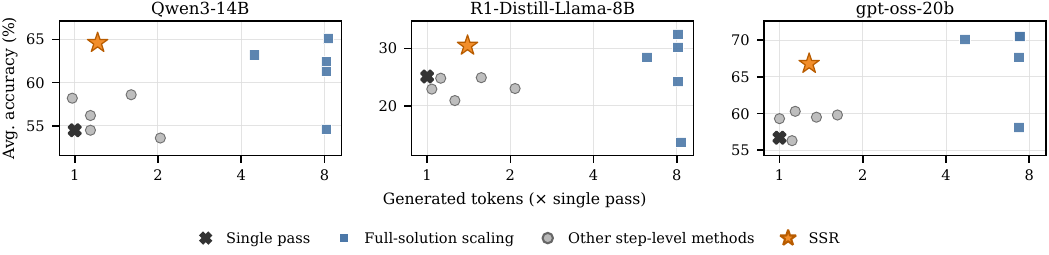}
    \captionsetup{skip=2pt}
    \caption{Macro-average accuracy across five benchmarks versus generated-token cost
    relative to single-pass decoding. Full-solution scaling includes
    Majority@8, ESC, ORM reranking, PRM reranking, and USC; other step-level
    methods include fixed-early resampling, low-confidence resampling, budget
    forcing, random rejection, and LATTS. \methodabbr{} achieves the highest
    accuracy among step-level methods for all three generators and remains
    competitive with full-solution scaling at substantially lower token cost.
    See Table~\ref{tab:end-to-end} for detailed results.}
    \label{fig:accuracy-token-tradeoff}
\end{figure*}

\section{Introduction}
\label{sec:introduction}
Reasoning language models benefit from additional inference-time computation,
but longer reasoning traces also create more opportunities to commit harmful
intermediate steps
\citep{wei2022chain,guo2025deepseek,muennighoff2025s1}.
Once a step is appended, subsequent generation is conditioned on it.
A model may recover through reflection, but it may instead elaborate a
plausible yet incorrect premise, reducing the chance of eventually reaching
the correct answer.

A natural response is to verify each step before committing it.
Process reward models score intermediate reasoning
\citep{lightman2024let,wang2024math}, and verifier-guided decoders can reject
low-scoring steps and sample alternatives \citep{uscidda2025latts}.
These approaches establish the utility of step-level intervention when a
capable verifier is available.
However, an external verifier may introduce additional model capacity or
supervision from stronger teachers, leaving open whether the generator itself
can learn to make these decisions.
We therefore ask: \emph{Given a reasoning prefix and one newly sampled step,
can a language model decide whether to commit that step or reject it and
sample again?}

We refer to this task as \emph{same-prefix candidate rejection}, which requires
distinguishing candidates sampled from the same problem and committed prefix.
Global verification metrics may reward recognition of problem difficulty,
trajectory depth, or accumulated errors without measuring this ability.
We therefore train and evaluate candidate discrimination within shared
prefixes.

A natural supervision signal is the change in \emph{recoverability}: the
probability that the frozen generator, continued from a prefix, eventually
produces a correct answer.
Yet adjacent changes in recoverability are difficult to estimate reliably
because they involve differences between noisy Monte Carlo success rates.
Across three reasoning models and five mathematical reasoning benchmarks,
only 2.3--24.0\% of adjacent recoverability differences are resolved with
16 rollouts per endpoint.
Cumulative degradation can nevertheless become detectable even when
individual transitions remain unresolved.
This motivates supervision based on cumulative degradation relative to the
problem's initial recoverability.

Cumulative degradation alone does not establish that resampling is useful.
We therefore also examine candidates sampled from shared prefixes.
These candidates exhibit variation beyond Monte Carlo noise, with a small
low-recoverability tail.
This suggests a selective intervention: accept most candidates immediately and
spend additional computation on replacing the few that are likely to harm the
continuation.

Our method, \method{} (\methodabbr), freezes the generator and trains a
lightweight LoRA \citep{hu2021lora} acceptance gate on the same backbone.
It uses confidence-qualified \emph{first-passage supervision} relative to the
problem's initial recoverability: confidently acceptable steps before the
first resolved crossing are labeled as accepted, the crossing step is labeled
as rejected, and unresolved steps and the subsequent suffix are excluded from
supervision.
Training combines pointwise classification with same-prefix pairwise learning,
directly encouraging the gate to distinguish accepted and rejected candidates
that share the same state.
We then refine the gate with a within-problem group-relative objective using
final-answer correctness.

At inference, the adapter is enabled only for acceptance scoring.
The decoder commits accepted candidates and resamples rejected ones from the
unchanged prefix, subject to a per-position attempt cap and an episode-level
rejection budget.
Additional sampling is therefore selective and requires no external learned
verifier.

Across three reasoning models and five mathematical reasoning benchmarks,
\methodabbr{} improves macro-average accuracy over single-pass decoding by
5.4--10.1 percentage points using 1.21--1.40$\times$ as many generated tokens.
It achieves the highest macro-average accuracy among the evaluated step-level
methods for each generator and remains competitive with full-solution scaling
methods, which use 4.47--8.27$\times$ the single-pass token cost.
Figure~\ref{fig:accuracy-token-tradeoff} summarizes this accuracy--cost
trade-off.

Our contributions are:
\begin{itemize}[leftmargin=*,nosep]
    \item We formulate \emph{same-prefix candidate rejection} and characterize
    its learning challenges: most adjacent recoverability changes are
    difficult to resolve at practical rollout budgets, while cumulative
    degradation is detectable and same-prefix candidates exhibit a sparse
    low-recoverability tail.

    \item We introduce \methodabbr{}, combining confidence-qualified
    first-passage supervision, same-prefix pairwise learning, and
    group-relative policy refinement to enable budgeted step rejection using
    the generator's own backbone.

    \item Through extensive experiments, we demonstrate \methodabbr{}'s
    favorable accuracy--token trade-off against both step-level and
    full-solution scaling baselines.
\end{itemize}

\section{Same-Prefix Candidate Rejection for Reasoning Models}
\label{sec:motivation}
\begin{figure*}[t]
    \centering
    \includegraphics[width=\textwidth]{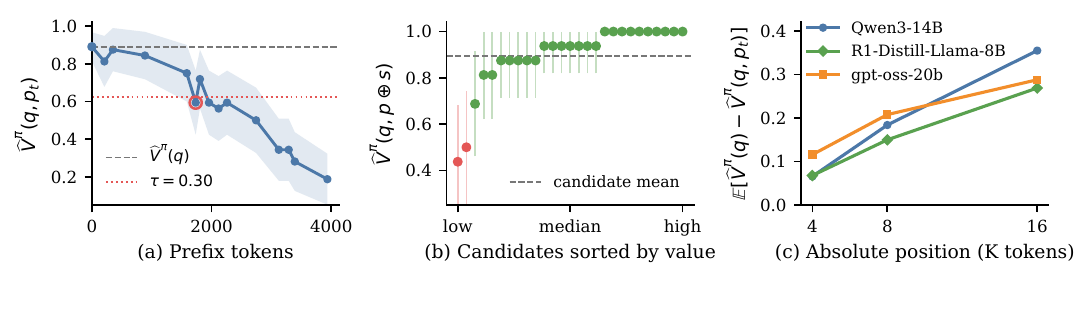}
    \captionsetup{skip=2pt}
    \caption{Why same-prefix candidate rejection is a distinct problem.
    (a) Recoverability along one trajectory evaluated at paragraph boundaries;
    the open circle marks the first crossing of a root-relative $\tau=0.30$
    barrier.
    (b) Candidate steps sampled from one shared prefix exhibit a small
    low-value tail.
    (c) Mean root-relative degradation increases with depth across the
    recoverability audit populations.}
    \label{fig:step-rejection-diagnostics}
\end{figure*}

\subsection{Problem Definition}

Let $q$ be a problem, $\pi$ a frozen reasoning policy, and
$p_t=(s_1,\ldots,s_t)$ a committed reasoning prefix.
At decision $t$, the policy samples
$s_t\sim\pi(\cdot\mid q,p_{t-1})$, and a gate either commits the candidate or
rejects it and resamples from the same conditional distribution.
The gate therefore receives $x_t=(q,p_{t-1},s_t)$.

For any prefix $p$, define its recoverability under the frozen policy as
\begin{equation}
V^\pi(q,p)
=
\Pr_{y\sim\pi(\cdot\mid q,p)}
\!\left[R(q,y)=1\right],
\label{eq:prefix-recoverability}
\end{equation}
where $R$ is a binary final-answer checker.
We write $V^\pi(q)=V^\pi(q,\varnothing)$ for the recoverability of the bare
problem and use $\widehat V^\pi$ for Monte Carlo estimates.
For a candidate step $s$, $p\oplus s$ denotes the extended prefix, so
$V^\pi(q,p\oplus s)$ is the probability of eventual success after committing
$s$ and continuing under $\pi$.

This same-prefix comparison distinguishes candidate rejection from global
verification.
A verifier that separates easy problems from hard ones, or scores only
properties accumulated across a trajectory, need not distinguish candidates
available at the current state.
Useful rejection instead requires
\begin{equation}
    \mathbb E[V^\pi(q,p\oplus s)\mid s\ \text{accepted}]
    >
    \mathbb E_{s\sim\pi(\cdot\mid q,p)}
    [V^\pi(q,p\oplus s)].
    \label{eq:accepted-value-condition}
\end{equation}
We therefore train and evaluate the gate within a shared prefix.

\subsection{Why Adjacent Recoverability Changes Are Difficult to Estimate}

We estimate recoverability from independent continuations.
A natural step-level target is the adjacent change
$\Delta_t=V^\pi(q,p_t)-V^\pi(q,p_{t-1})$, but it is the difference between two
noisy Bernoulli means.

\begin{figure}[t]
    \centering
    \includegraphics[width=\linewidth]{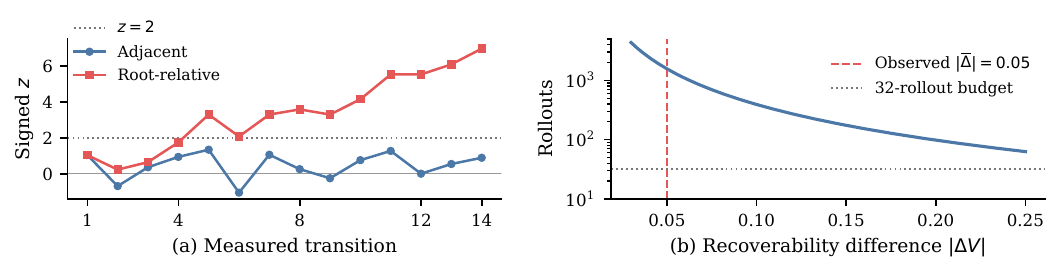}
    \captionsetup{skip=2pt}
    \caption{Cumulative degradation is statistically visible when
    local degradation is not.
    (a) Signed likelihood-ratio statistics for adjacent and root-relative
    changes along the measured trajectory. Positive values indicate a loss of
    recoverability; the dotted line at $z=2$ provides an approximate
    significance reference.
    (b) Approximate rollouts per endpoint required to distinguish two
    Bernoulli success probabilities with 80\% power.}
    \label{fig:value-difference-power}
\end{figure}

Figure~\ref{fig:step-rejection-diagnostics}(a) shows one Qwen3-14B trajectory
evaluated at paragraph boundaries.
Estimated recoverability falls from $0.891$ to $0.188$, with temporary
increases and no statistically resolved adjacent transition
(Figure~\ref{fig:value-difference-power}(a)).
The mean adjacent change is approximately $-0.05$; a standard power calculation
requires roughly 1,600 rollouts per endpoint to resolve an effect of this size
with 80\% power (Figure~\ref{fig:value-difference-power}(b)).
Appendix~\ref{sec:diagnostic-cases} provides the power calculation.

Although individual changes are weak, their cumulative effect is measurable.
Define the root-relative degradation
\begin{equation}
    D_t=V^\pi(q)-V^\pi(q,p_t).
\label{eq:cumulative-drop}
\end{equation}
The trajectory in Figure~\ref{fig:step-rejection-diagnostics}(a) reaches
$D_t=0.703$, and its first crossing of a fixed root-relative barrier is clearly
observable.
Figure~\ref{fig:step-rejection-diagnostics}(c) shows the same increasing
degradation with depth across the broader audit populations.
The root serves as a difficulty-normalized reference rather than an assumed
maximum.
These observations motivate supervising the first measurable cumulative
crossing rather than regressing noisy adjacent value changes.

\subsection{Candidate Values Exhibit a Rare Bad Tail}

Trajectory-level degradation alone does not imply that resampling from a
fixed prefix is useful.
We therefore sample candidate openings from a shared prefix and estimate their
recoverability using independent continuations.
Figure~\ref{fig:step-rejection-diagnostics}(b) reports 27 usable candidates,
each evaluated with 16 continuations.
Their variation exceeds the estimated Monte Carlo noise, indicating
within-prefix differences.
Most candidates retain high recoverability, with a median value of $0.938$
and only two candidates at or below $0.5$.
This sparse low-value tail motivates selectively rejecting candidates while
accepting most candidates without additional sampling.
Together with measurable cumulative degradation, this supports training the
acceptance gate with root-relative first-passage labels.

\section{Self-Step Rejection}
\label{sec:method}
\begin{figure*}[t]
    \centering
    \includegraphics[width=\textwidth]{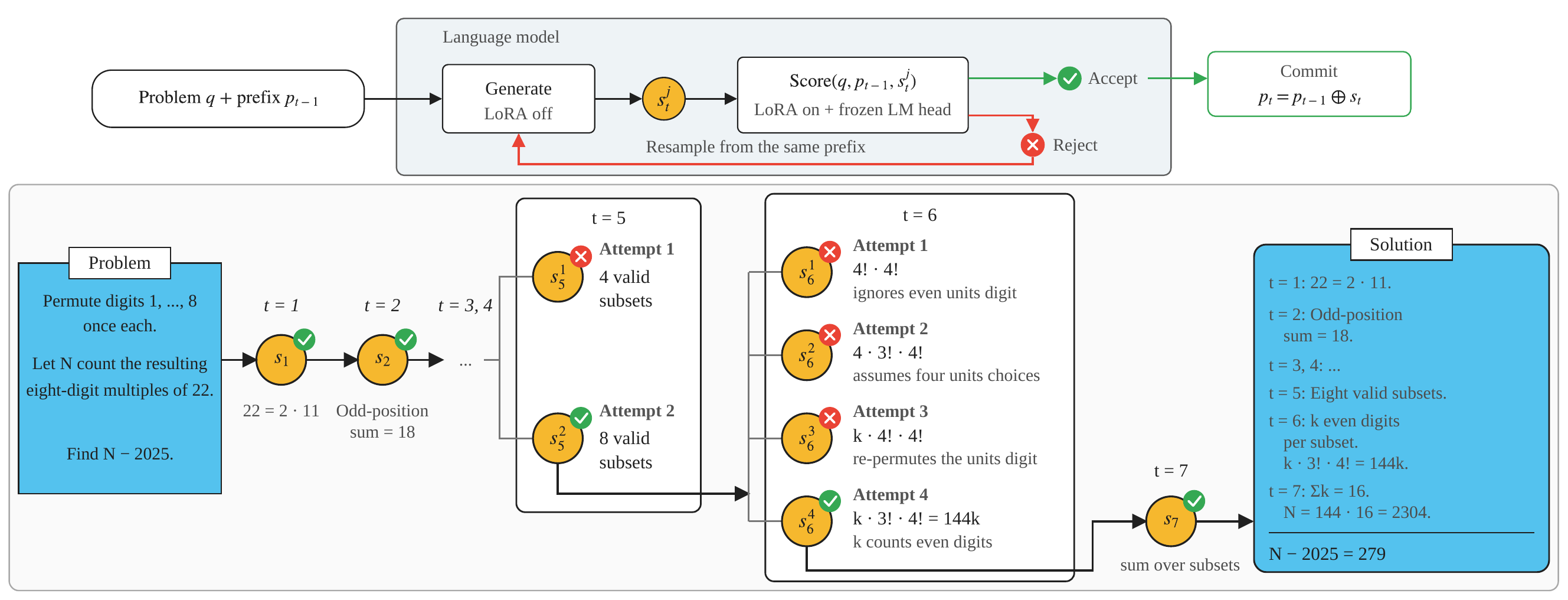}
    \captionsetup{skip=6pt}
    \caption{Top: \methodabbr{} uses the frozen model to generate candidates
    and enables LoRA to score them.
    Acceptance extends the prefix; rejection triggers resampling from the
    unchanged prefix.
    Bottom: an actual decoding example with at most $L_{\mathrm{step}}$ new
    tokens per step, shown as reasoning summaries. Attempts at each position proceed
    from top to bottom.
    Resampling is bounded by a per-position attempt cap $M$ and an
    episode-level rejection budget $B$.}
    \label{fig:ssr-inference}
\end{figure*}

\method{} (\methodabbr) trains a same-model acceptance gate through
first-passage supervision, same-prefix pairwise learning, and group-relative
policy refinement, then deploys it for budgeted accept-or-resample decoding.
Figure~\ref{fig:ssr-inference} illustrates the sparse inference-time
intervention and the same-model generate--judge loop.

\subsection{First-Passage Supervision}

For each training problem, we sample one or more trajectories from the frozen
generator $\pi$ one step at a time, with each step containing at most
$L_{\mathrm{step}}$ newly generated tokens.
Let $\widehat V_0=\widehat V^\pi(q)$ be the root estimate and
$\widehat V_t=\widehat V^\pi(q,p_t)$ the estimate after committing step $t$.
Ignoring estimation uncertainty, the first root-relative crossing for a
degradation threshold $\tau$ is
\begin{equation}
    t^\star
    =
    \min\left\{
    t:
    \widehat V_t\le\widehat V_0(1-\tau)
    \right\}.
\label{eq:first-crossing}
\end{equation}
If no step crosses the barrier, $t^\star=\infty$.

In practice, labeling stops at the first confidence-resolved crossing, which
may occur later than the point-estimate crossing $t^\star$.
Confidence intervals account for Monte Carlo uncertainty in the difference
$\widehat V_t-(1-\tau)\widehat V_0$.
Preceding steps confidently above the barrier are labeled \textsc{accept},
and the crossing step is labeled \textsc{reject}.
The subsequent suffix remains unlabeled because, after the first detected
barrier crossing, low recoverability may be inherited rather than caused by
each subsequent step.
Additional rollouts are allocated adaptively near the boundary; steps that
remain unresolved are excluded from supervision.
Appendix~\ref{sec:ssr-appendix} gives the complete labeling procedure and edge
cases.

\subsection{Policy-Matched Acceptance Scoring}

For a candidate step $s_t$, we form the gate input
$x_t=(q,p_{t-1},s_t)$.
For scoring, we serialize $x_t$, append a fixed reserved marker, and run the
same backbone with a LoRA adapter \citep{hu2021lora} $\Delta_\phi$ enabled.
Following token-based verifier readouts \citep{wang2024math}, let
$\ell_\phi^{+}(x_t)$ and $\ell_\phi^{-}(x_t)$ denote the frozen
language-modeling head's logits at the marker position for the existing
vocabulary tokens \texttt{+} and \texttt{-}, respectively.
We define the acceptance logit and normalized acceptance score as
\begin{equation}
    z_\phi(x_t)
    =
    \ell_\phi^{+}(x_t)-\ell_\phi^{-}(x_t),
    \qquad
    \operatorname{score}_\phi(x_t)
    =
    \frac{\exp\ell_\phi^{+}(x_t)}
    {\exp\ell_\phi^{+}(x_t)+\exp\ell_\phi^{-}(x_t)}
    =
    \sigma\!\left(z_\phi(x_t)\right).
\label{eq:acceptance-score}
\end{equation}
The score represents acceptance confidence rather than an estimate of
recoverability.
The base parameters $\theta$, marker embedding, and language-modeling head
remain frozen; only the LoRA adapter is trained.
The adapter is disabled during candidate generation and enabled only for
acceptance scoring.

\subsection{Same-Prefix Supervised Objective}

Let $\mathcal D$ contain the confidence-qualified examples, with $y_t=1$ for
\textsc{accept} and $y_t=0$ for \textsc{reject}.
The pointwise objective is weighted binary cross entropy:
\begin{equation}
\mathcal L_{\mathrm{BCE}}
=
-\sum_{(x_t,y_t)\in\mathcal D}
\omega_{y_t}
\left[
y_t\log \operatorname{score}_\phi(x_t)
+
(1-y_t)\log(1-\operatorname{score}_\phi(x_t))
\right],
\label{eq:bce-loss}
\end{equation}
where $\omega_{y_t}$ is a class-dependent optimization weight.
In implementation, this objective is evaluated with binary cross entropy on
the logit difference $z_\phi(x_t)$ for numerical stability.

Pointwise accuracy can be high while a gate learns only problem
difficulty or trajectory depth.
We therefore enrich states near the crossing with multiple same-prefix candidates
$s_t^1,\ldots,s_t^M$ sampled from the same prefix.
Let $\mathcal P_t$ contain the confidence-qualified pairs $(i,j)$ from this
group, where candidate $i$ is accepted and candidate $j$ is rejected.
Following RankNet \citep{burges2005learning}, we add the pairwise objective
\begin{equation}
\mathcal L_{\mathrm{pair}}
=
\sum_{(i,j)\in\mathcal P_t}
\log\left(
1+\exp[-(z_i-z_j)]
\right),
\qquad
z_i=z_\phi(x_t^i).
\label{eq:same-prefix-loss}
\end{equation}
The complete supervised objective is
\begin{equation}
    \mathcal L_{\mathrm{gate}}
    =
    \mathcal L_{\mathrm{BCE}}
    +
    \lambda_{\mathrm{pair}}\mathcal L_{\mathrm{pair}}.
\label{eq:ssr-supervised-loss}
\end{equation}
This objective directly rewards within-prefix ordering, reducing reliance on
problem difficulty or trajectory depth alone.

\subsection{Group-Relative Policy Refinement}

We refine the supervised gate to improve final-answer accuracy under budgeted
sequential decoding using a REINFORCE-style objective \citep{williams1992simple}
with a group-relative baseline \citep{shao2024deepseekmath}.
We optimize the gate as a stochastic binary policy
$\rho_\phi(u_t\mid x_t)$ over
$u_t\in\{\textsc{accept},\textsc{reject}\}$:
\begin{equation}
    \rho_\phi(\textsc{accept}\mid x_t)
    =
    \sigma\!\left(
        \frac{z_\phi(x_t)-\operatorname{logit}(\eta)}
        {\mathrm{temperature}}
    \right),
    \qquad
    \rho_\phi(\textsc{reject}\mid x_t)
    =
    1-\rho_\phi(\textsc{accept}\mid x_t),
\label{eq:rl-policy}
\end{equation}
where $\mathrm{temperature}$ controls action sampling and
$\eta$ is fixed during policy refinement and deployment.
Training samples actions from $\rho_\phi$; inference uses the deterministic
threshold rule $\operatorname{score}_\phi(x_t)\ge\eta$.

For each problem, we sample a group of $G$ complete decoding episodes.
Let $R_e\in\{0,1\}$ denote final-answer correctness for episode $e$ and define
the within-problem advantage
\begin{equation}
    A_e
    =
    R_e-\frac{1}{G}\sum_{e'=1}^{G}R_{e'}.
\label{eq:group-relative-advantage}
\end{equation}
The group baseline compares each episode's outcome with other episodes of the
same problem.
With $\mathcal T_e$ denoting the gate decisions in episode $e$, we optimize
\begin{equation}
    \mathcal L_{\mathrm{RL}}
    =
    -\frac{1}{\sum_e|\mathcal T_e|}
    \sum_{e=1}^{G}
    A_e
    \sum_{t\in\mathcal T_e}
    \log\rho_\phi(u_{e,t}\mid x_{e,t}).
\label{eq:group-relative-loss}
\end{equation}
The environment enforces a per-position attempt cap and an episode-level
rejection budget, bounding generation cost while the terminal reward optimizes
correctness.

\subsection{Sequential Accept-or-Resample Decoding}

At inference time, the base model samples one candidate at a time.
The acceptance gate commits the candidate when
$\operatorname{score}_\phi(x_t)\ge\eta$.
Each position receives at most $M$ candidate attempts.
Rejected candidates are discarded and resampled from the unchanged parent
prefix until either a candidate is accepted or the attempt cap is reached; in
the latter case, the decoder commits the highest-scoring attempt.
Across the episode, at most $B$ candidates may be rejected.
Once this budget is exhausted, subsequent candidates are committed directly.
\methodabbr{} generates additional candidates only after a rejection,
allocating extra computation selectively within the prescribed budgets.
Step-level best-of-$K$, in comparison, incurs the cost of sampling $K$
candidates at every decision.

\section{Experiments}
\label{sec:experiment}
\begin{table*}[t]
\centering
\scriptsize
\setlength{\tabcolsep}{3pt}
\renewcommand{\arraystretch}{0.94}
\caption{End-to-end accuracy (\%) and generation cost, averaged over
three evaluation runs. ``Avg.'' is the macro-average and
``Tok.$\times$'' is relative to single-pass generation. Gray rows are
step-level methods; underlining and bold mark the best full-solution-scaling and
step-level accuracy, respectively.}
\label{tab:end-to-end}
\begin{tabular}{llccccccc}
\toprule
Model & Method & A25 & Olym-E & Olym-H & HMMT25 & HMMT26
& Avg. & Tok.$\times$ \\
\midrule
Qwen3-14B
& Single pass        & 70.0 & 83.0 & 21.0 & 50.0 & 48.5 & 54.5 & 1.00 \\
& Majority@8         & 76.7 & 94.0 & 33.0 & 56.7 & 51.5 & 62.4 & 8.11 \\
& ESC ($w=2$)        & 77.8 & \underline{94.7} & 34.3 & 56.7 & 52.5 & 63.2 & 4.47 \\
& ORM reranking      & 76.7 & 91.0 & 28.0 & 53.3 & \underline{57.6} & 61.3 & 8.11 \\
& PRM reranking      & 70.0 & 85.0 & 23.0 & 46.7 & 48.5 & 54.6 & 8.11 \\
& USC                & \underline{80.0} & 93.0 & \underline{35.0} & \underline{60.0} & \underline{57.6} & \underline{65.1} & 8.24 \\
\rowcolor{gray!12}\cellcolor{white}
& Fixed early        & 73.3 & 90.0 & 28.0 & 53.3 & 48.5 & 58.6 & 1.60 \\
\rowcolor{gray!12}\cellcolor{white}
& Low-conf. resample & 73.3 & 87.0 & 22.0 & \textbf{63.3} & 45.5 & 58.2 & 0.98 \\
\rowcolor{gray!12}\cellcolor{white}
& Budget forcing     & 66.7 & 81.0 & 22.0 & 50.0 & 48.5 & 53.6 & 2.04 \\
\rowcolor{gray!12}\cellcolor{white}
& Random reject      & 66.7 & 86.7 & 23.3 & 53.3 & 42.4 & 54.5 & 1.14 \\
\rowcolor{gray!12}\cellcolor{white}
& LATTS              & 70.0 & 80.0 & 33.3 & 43.3 & \textbf{54.5} & 56.2 & 1.14 \\
\rowcolor{gray!12}\cellcolor{white}
& \methodabbr{}      & \textbf{77.8} & \textbf{95.7} & \textbf{35.0} & 62.2 & 52.5 & \textbf{64.6} & 1.21 \\
\midrule
R1-Distill-Llama-8B
& Single pass        & 33.3 & 36.0 &  5.0 & 30.0 & 21.2 & 25.1 & 1.00 \\
& Majority@8         & \underline{40.0} & \underline{48.0} &  6.0 & 26.7 & 30.3 & 30.2 & 8.06 \\
& ESC ($w=2$)        & 35.6 & 44.0 &  7.0 & 25.6 & 30.3 & 28.5 & 6.23 \\
& ORM reranking      & \underline{40.0} & 46.0 & \underline{10.0} & \underline{30.0} & \underline{36.4} & \underline{32.5} & 8.06 \\
& PRM reranking      & 30.0 & 37.0 &  7.0 & 16.7 & 30.3 & 24.2 & 8.06 \\
& USC                & 10.0 & 31.0 &  5.0 & 10.0 & 12.1 & 13.6 & 8.27 \\
\rowcolor{gray!12}\cellcolor{white}
& Fixed early        & \textbf{40.0} & 35.0 &  5.0 & 23.3 & 21.2 & 24.9 & 1.57 \\
\rowcolor{gray!12}\cellcolor{white}
& Low-conf. resample & 26.7 & 36.7 &  3.3 & 26.7 & 21.2 & 22.9 & 1.04 \\
\rowcolor{gray!12}\cellcolor{white}
& Budget forcing     & 33.3 & 33.3 &  3.3 & 26.7 & 18.2 & 23.0 & 2.08 \\
\rowcolor{gray!12}\cellcolor{white}
& Random reject      & 20.0 & 40.0 &  6.7 & 16.7 & 21.2 & 20.9 & 1.26 \\
\rowcolor{gray!12}\cellcolor{white}
& LATTS              & 30.0 & 33.3 & \textbf{16.7} & 20.0 & 24.2 & 24.8 & 1.12 \\
\rowcolor{gray!12}\cellcolor{white}
& \methodabbr{}      & 37.8 & \textbf{42.7} & 10.3 & \textbf{32.2} & \textbf{29.3} & \textbf{30.5} & 1.40 \\
\midrule
gpt-oss-20b
& Single pass        & 70.0 & 79.0 & 27.0 & 50.0 & 57.6 & 56.7 & 1.00 \\
& Majority@8         & \underline{86.7} & \underline{93.0} & 37.0 & 66.7 & 69.7 & \underline{70.6} & 7.34 \\
& ESC ($w=2$)        & \underline{86.7} & 91.3 & 34.0 & 67.8 & \underline{70.7} & 70.1 & 4.68 \\
& ORM reranking      & 73.3 & 87.0 & \underline{42.0} & 66.7 & 69.7 & 67.7 & 7.34 \\
& PRM reranking      & 70.0 & 76.0 & 30.0 & 56.7 & 57.6 & 58.1 & 7.34 \\
& USC                & \underline{86.7} & 89.0 & 37.0 & \underline{70.0} & 69.7 & 70.5 & 7.40 \\
\rowcolor{gray!12}\cellcolor{white}
& Fixed early        & 76.7 & 80.0 & 34.0 & 56.7 & 51.5 & 59.8 & 1.62 \\
\rowcolor{gray!12}\cellcolor{white}
& Low-conf. resample & 76.7 & 81.0 & 28.0 & 53.3 & 57.6 & 59.3 & 1.00 \\
\rowcolor{gray!12}\cellcolor{white}
& Budget forcing     & 73.3 & 80.0 & 33.0 & 56.7 & 54.5 & 59.5 & 1.36 \\
\rowcolor{gray!12}\cellcolor{white}
& Random reject      & \textbf{80.0} & 80.0 & 33.3 & 56.7 & 51.5 & 60.3 & 1.14 \\
\rowcolor{gray!12}\cellcolor{white}
& LATTS              & 66.7 & 70.0 & 33.3 & \textbf{60.0} & 51.5 & 56.3 & 1.11 \\
\rowcolor{gray!12}\cellcolor{white}
& \methodabbr{}      & 78.9 & \textbf{81.7} & \textbf{45.7} & 58.9 & \textbf{68.7} & \textbf{66.8} & 1.28 \\
\bottomrule
\end{tabular}
\end{table*}

\subsection{Setup}

We study Qwen3-14B \citep{yang2025qwen3},
DeepSeek-R1-Distill-Llama-8B \citep{guo2025deepseek}, and
gpt-oss-20b \citep{agarwal2025gpt} as frozen reasoning generators.
Our primary evaluation suite contains contest-mathematics problems from
OlymMATH-Easy and OlymMATH-Hard \citep{sun2026challenging}, together with
AIME 2025, HMMT February 2025, and HMMT February 2026 from
MathArena \citep{dekoninck2026beyond}.
We train the acceptance gate on a filtered subset of
Omni-MATH \citep{gao2025omni}, decontaminated against all evaluation
benchmarks, and reserve a separate subset for development.

We compare \methodabbr{} with the following baselines.
Single pass is the no-scaling reference, while
Majority@8 \citep{wang2022self} and ESC ($w=2$) \citep{li2024escape}
represent fixed-budget and adaptive self-consistency.
ORM reranking \citep{liu2026skywork}, PRM reranking \citep{zhang2025lessons},
and USC \citep{chen2023universal} select among eight complete solutions.
We design fixed-early resampling, low-confidence resampling, and random
rejection as controls for intervention timing and candidate selection.
Budget forcing \citep{muennighoff2025s1} tests the effect
of extending generation.
LATTS \citep{uscidda2025latts} provides an external-verifier step-rejection baseline.
Fixed early and random rejection use the same generation-budget constraints as
\methodabbr{}.

\subsection{Implementation Details}

Unless otherwise specified, all generators use a 32K context window and a per-step limit of
$L_{\mathrm{step}}=2048$ new tokens, reduced when context space is insufficient.
Each generator has a separate rank-16 LoRA gate on the query, output, and MLP
projections, with key and value projections unchanged.
The adapter is enabled only for scoring, using the frozen
language-modeling head's \texttt{+}/\texttt{-} logits at a fixed marker.
Training combines first-passage supervision with group-relative policy
refinement.
Inference allows $M=8$ candidate attempts per position and $B=12$ rejections
per episode.
A threshold $\eta$, calibrated before refinement to a target rejection rate
of 17.8\%, remains fixed during refinement and inference.
Appendix~\ref{sec:implementation-details} provides decoding, optimization,
calibration, and runtime details.

\begin{table*}[t]
\centering
\small
\setlength{\tabcolsep}{4pt}
\caption{Recoverability diagnostics across five benchmarks.
$\mathbb E[|\Delta_t|]$: mean absolute adjacent change;
resolved@16: fraction resolved with 16 rollouts per endpoint;
Cand.\ SD: noise-corrected within-prefix candidate standard deviation;
Tail: fraction significantly below the best candidate.
Brackets indicate 95\% problem-bootstrap confidence intervals.}
\label{tab:problem-characterization}
\resizebox{0.85\textwidth}{!}{%
\begin{tabular}{lcccc}
\toprule
Model & $\mathbb E[|\Delta_t|]$ & Adj.\ resolved@16 (\%) & Cand.\ SD & Tail (\%) \\
\midrule
Qwen3-14B
& 0.215 {\scriptsize [0.175, 0.253]} & 24.0 {\scriptsize [19.2, 28.2]}
& 0.113 {\scriptsize [0.027, 0.176]} & 6.0 {\scriptsize [2.0, 12.0]} \\
R1-Distill-Llama-8B
& 0.051 {\scriptsize [0.014, 0.097]} & 2.3 {\scriptsize [0.0, 9.2]}
& 0.120 {\scriptsize [0.058, 0.168]} & 11.0 {\scriptsize [3.0, 21.0]} \\
gpt-oss-20b
& 0.140 {\scriptsize [0.071, 0.227]} & 18.7 {\scriptsize [2.9, 47.2]}
& 0.184 {\scriptsize [0.067, 0.265]} & 11.5 {\scriptsize [3.1, 21.9]} \\
\bottomrule
\end{tabular}
}
\end{table*}

\begin{figure*}[t]
\centering
\includegraphics[width=\textwidth]{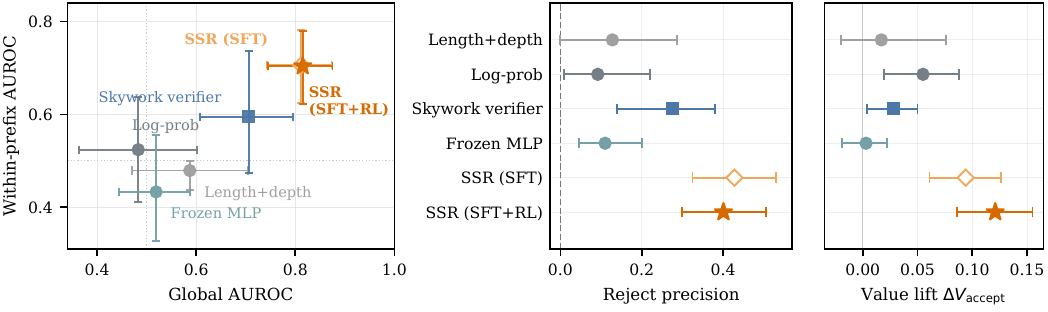}
\captionsetup{skip=2pt}
\caption{Left: verifier discrimination measured by global AUROC over pooled
candidates and within-prefix AUROC over candidates sampled from the same
prefix; error bars are 95\% confidence intervals. Center: reject
precision at a fixed rejection rate of 17.8\%. Right: accepted-candidate value lift.
Qwen3-14B is the generator throughout.
Skywork verifier denotes Skywork-Reward-V2-Qwen3-8B \citep{liu2026skywork}.}
\label{fig:classifier-diagnostics}
\end{figure*}

\begin{table}[t]
\centering
\scriptsize
\setlength{\tabcolsep}{4pt}
\caption{Training-component ablations on Qwen3-14B, averaged over three
evaluation runs. Indented rows remove or replace one component of the full
\methodabbr{} recipe. ``Avg.'' is the five-benchmark macro-average and
``Tok.$\times$'' is relative to single-pass generation.}
\label{tab:training-ablation}
\begin{tabular}{lcccccccc}
\toprule
Method & A25 & Olym-E & Olym-H & HMMT25 & HMMT26 & Avg. & Tok.$\times$ \\
\midrule
Single pass
& 70.0 & 83.0 & 21.0 & 50.0 & 48.5 & 54.5 & 1.00 \\
\midrule
\methodabbr{} (full)
& \textbf{77.8} & \textbf{95.7} & \textbf{35.0} & \textbf{62.2}
& \textbf{52.5} & \textbf{64.6} & 1.21 \\
\quad Label all below-barrier steps
& 72.2 & 86.0 & 24.3 & 53.3 & 49.5 & 57.1 & 1.62 \\
\quad No same-prefix pairwise loss
& 73.3 & 88.0 & 26.3 & 54.4 & 50.5 & 58.5 & 1.14 \\
\quad No policy refinement
& 74.4 & 89.7 & 28.7 & 56.7 & 50.5 & 60.0 & 1.12 \\
\bottomrule
\end{tabular}
\end{table}

\subsection{Main Results}

Table~\ref{tab:end-to-end} shows that \methodabbr{} improves macro-average
accuracy over single-pass decoding by 5.4--10.1 percentage points using
1.21--1.40$\times$ as many generated tokens, with gains across all 15
model--benchmark pairs.
The learned gate distinguishes same-prefix candidates to reject low-value
continuations before committing them.
\methodabbr{}'s advantage over random rejection and LATTS at comparable token
costs suggests that candidate discrimination matters beyond resampling alone,
consistent with the diagnostics and ablations below.
By resampling only after rejection and preserving the accepted prefix,
\methodabbr{} concentrates extra generation on local alternatives.
This selective allocation helps explain its strong accuracy--cost trade-off:
\methodabbr{} achieves the highest step-level macro-average accuracy for all
three generators and remains competitive with full-solution scaling.
The strongest full-solution baseline for each generator achieves 0.5--3.8
percentage points higher accuracy but uses 7.34--8.24$\times$ the single-pass
token cost.
Figure~\ref{fig:accuracy-token-tradeoff} summarizes this trade-off.

\subsection{Analysis}

\paragraph{Problem characterization.}
Table~\ref{tab:problem-characterization} shows that only 2.3--24.0\% of
adjacent differences are resolved with 16 rollouts per endpoint, despite
nonzero within-prefix variation across all three models.
This suggests that noisy local estimates can obscure useful candidate
differences, motivating supervision based on cumulative degradation.
Meanwhile, only 6.0--11.5\% of candidates occupy the low-value tail,
supporting selective rejection while preserving most of the trajectory.

\paragraph{Same-prefix gate analysis.}
Figure~\ref{fig:classifier-diagnostics} shows that strong global discrimination
does not necessarily translate into effective within-prefix selection, where
problem difficulty and prefix depth cannot distinguish candidates.
\methodabbr{}'s stronger local discrimination is consistent with its pairwise
training objective.
Policy refinement raises accepted-candidate value lift from 0.094 to 0.121
with modest changes in discrimination metrics, suggesting improvements in
acceptance decisions beyond offline ranking.

\paragraph{Training-component ablations.}
Table~\ref{tab:training-ablation} shows that labeling all below-barrier steps
reduces accuracy by 7.5 points and increases token cost, consistent with
unnecessary rejection of steps that inherit earlier degradation.
Removing pairwise learning costs 6.1 points, supporting direct supervision
of local candidate preferences.
Omitting policy refinement costs 4.6 points, suggesting that candidate
discrimination benefits from further optimization for final-answer correctness
under sequential decoding.

\paragraph{Sensitivity analysis.}
On Qwen3-14B (Appendix~\ref{sec:additional-results},
Tables~\ref{tab:rejection-rate-sensitivity} and~\ref{tab:step-sensitivity}),
accuracy plateaus at target rejection rates of 17.8--30.0\% while token cost
rises, indicating diminishing returns from additional rejection.
Step sizes of 1,024--4,096 tokens achieve similar accuracy, whereas shorter
and longer steps perform worse.
Short steps increase gate decisions, while long steps delay intervention,
suggesting a balance between decision frequency and intervention timing.
The default $L_{\mathrm{step}}=2048$ uses the fewest tokens within this
accuracy plateau.

\section{Related Work}
\label{sec:related_work}
\paragraph{Inference-time scaling and search.}
Sampling multiple complete solutions and aggregating their answers improves
reasoning robustness \citep{wang2022self,cobbe2021training}.
Tree-structured and agentic methods instead branch at intermediate states and use search,
learned values, or structured subroutines to allocate computation
\citep{yao2023tree,luo2024improve,rma,proto,zhao2021augmenting}.
Adaptive methods control different aspects of this computation:
ESC adjusts the number of sampled solutions through early stopping
\citep{li2024escape}, while s1 uses budget forcing to control the length of a
single reasoning trace \citep{muennighoff2025s1}.
The effectiveness of scaling strategies also depends on problem difficulty
and context quality, motivating adaptive allocation or selection tailored to
individual problems
\citep{snell2024scaling,sun2026learningtrustselectivecontext}.
Multi-agent coordination and collaborative agent frameworks provide
other ways to guide reasoning
\citep{zhang2026agentrouter,zhang2026mapro,huang2026agent}.
We study allocation within a single decoding trajectory: the gate decides
which candidates warrant replacement before commitment, concentrating extra
generation on selected positions while preserving the accepted prefix.

\paragraph{Process supervision and verifier-guided decoding.}
Process reward models assign scores to intermediate reasoning steps
\citep{lightman2024let,wang2024math}.
Rollout-based supervision estimates successful completion from a prefix,
reducing manual annotation while depending on the rollout policy and sample count
\citep{xiong2026enhancing,xiong2025enhancing}.
OmegaPRM lowers labeling cost by searching for the first error under an
approximate monotonicity assumption \citep{luo2024improve}, while Process
Advantage Verifiers assess action-induced progress from a shared state
\citep{setlur2025rewarding}.
Recoverability and calibrated uncertainty have also been used to guide policy optimization
\citep{bao2025conflict,zhang2026optimizing,zhang2026semantic}.
For inference-time selection, Tree-PLV learns from same-prefix rollout
preferences to guide tree search \citep{he2024advancing}, whereas LATTS uses an
external PRM or LLM critic to reject steps and trigger resampling
\citep{uscidda2025latts}.
Our measurements motivate first-passage supervision when adjacent value
changes are difficult to resolve but cumulative degradation is detectable.
Building on verifier-guided decoding, we focus on learning a policy-matched
acceptance gate from first-passage supervision and evaluating its ability to
distinguish same-prefix candidates.

\paragraph{Self-evaluation and self-verification.}
Model representations and generated reasoning traces can expose signals useful
for verification and responsible reasoning evaluation
\citep{burns2022discovering,sun2026accuracymeasuringbiasacknowledgment}.
Self-Refine uses the same model to generate feedback and iteratively revise
its output without additional training \citep{madaan2023self},
although intrinsic self-correction without external feedback can fail to
improve reasoning \citep{huang2024large}.
Training-based approaches include SCoRe, which learns answer revision through
multi-turn reinforcement learning \citep{kumar2025training}, and
S$^2$R, which trains explicit self-verification and correction behaviors
\citep{ma2025s2r}.
Related work also explores reflection, self-evolving skills, and agent--environment
co-evolution \citep{wang2026reflex,wang2026atlasva,hu2026seal}.
Logical grounding and structure-based validation are studied in related
reasoning and planning settings
\citep{xiong2025deliberate,bao2026rlearner,bao2026scitrue,lu2025uni,lu2026mllms,ye2026osef,li2026gracegroundedreasoningadapter}.
\method{} instead formulates same-prefix candidate rejection: a learned gate
evaluates a candidate before commitment and, after rejection, resamples from
the unchanged prefix.
We introduce first-passage supervision for this acceptance gate, trained as a
lightweight adapter while keeping the generator frozen.

\section{Conclusion}
\label{sec:conclusion}
We presented \method{}, which trains a lightweight acceptance gate on the
generator's own backbone to reject low-value candidate steps without an
external learned verifier.
The method combines root-relative first-passage supervision, same-prefix
pairwise learning, and policy refinement while keeping the generator frozen.
Across three reasoning models and five mathematics benchmarks, \methodabbr{}
improves macro-average accuracy over single-pass decoding by 5.4--10.1
percentage points using 1.21--1.40$\times$ as many generated tokens,
achieving the highest macro-average accuracy among evaluated step-level
methods for each generator.
These results demonstrate the value of learning same-prefix candidate
discrimination to guide selective resampling and improve the
accuracy--token trade-off.

\subsection*{AI use statement}
We used generative AI tools to assist with writing and language
editing during manuscript preparation. The authors reviewed and
revised the AI-assisted text and take full responsibility for
the final manuscript.

\bibliography{iclr2027_conference}
\bibliographystyle{iclr2027_conference}

\newpage
\appendix

\label{sec:appendix}
\section{Self-Step Rejection}
\label{sec:ssr-appendix}

\subsection{First-Passage Labeling}
\label{sec:first-passage-labeling}

First-passage labeling converts trajectory-level value estimates into
step-level accept/reject supervision.
For a trajectory $p_0,\ldots,p_T$, let
$\widehat V_{\mathrm{barrier}}=(1-\tau)\widehat V_0$ denote the root-relative
barrier and define
$\mathrm{gap}_t=\widehat V_{\mathrm{barrier}}-\widehat V_t$.
The first confidence-resolved step with $\mathrm{gap}_t>0$ is labeled
\textsc{reject}; all preceding confidence-resolved steps are labeled
\textsc{accept}.
The procedure then terminates and assigns no labels to the remaining suffix.
This restriction associates the negative label with the first
confidence-resolved crossing of the root-relative barrier, avoiding repeated
attribution of inherited damage to later steps.
We evaluate dense labeling of every below-barrier step as an ablation.

At each prefix, the rollout count is increased only while the confidence
interval for $\mathrm{gap}_t$ intersects zero.
An interval lying strictly below zero yields an \textsc{accept} label, whereas
an interval lying strictly above zero yields the first \textsc{reject} label
and terminates the scan.
Prefixes that remain unresolved at $K_{\max}$ are marked gray and excluded from
the supervised objective.
This abstention rule avoids converting Monte Carlo ties into arbitrary labels.

If $\widehat V_0=0$, the relative barrier is also zero and contains no
information about $\tau$; Algorithm~\ref{alg:first-passage-labeling} therefore
returns no labels.
We likewise exclude candidate sets for which every evaluated continuation has
zero value, because such sets contain no within-prefix signal for selecting one
step over another.
Labels are taken from the confidence-qualified decisions recorded during data
collection rather than recomputed from the stored value estimates.
Adaptive stopping changes the conditional distribution of the stored
estimates, but not the recorded decision or committed candidate.
The resulting first-passage data are positive-heavy, so we report an
always-accept gate as the explicit no-intervention baseline.

The relative barrier scales the tolerated degradation with each problem's
recoverability.
A fixed absolute margin cannot trigger when the root estimate is below that
margin.
The root estimate is shared by all trajectories of a problem and can be
computed once.

\subsection{Group-Relative Policy Refinement}
\label{sec:rl-details}

Policy refinement samples $G$ complete decoding episodes for each problem and
uses final-answer correctness as the terminal reward.
Gate actions are sampled according to Equation~\ref{eq:rl-policy}.
The within-problem group mean provides a parameter-free baseline that reduces
the influence of problem difficulty without requiring a learned value function.
Rewards are terminal and returns are undiscounted, so all decisions in an
episode share the group-centered advantage in
Equation~\ref{eq:group-relative-advantage}.

We center rewards by the group mean without standardizing by the group standard
deviation.
For binary rewards, the standard deviation is fully determined by the empirical
success rate, $\sqrt{\widehat\mu(1-\widehat\mu)}$.
With $G=8$, standardization would give a group with
$\widehat\mu=1/8$ approximately $1.5\times$ the group-level scaling of a
balanced group with $\widehat\mu=4/8$, emphasizing a near-degenerate group
whose baseline is estimated from a single successful episode.
Mean centering preserves the observed reward scale, and constant-outcome groups
naturally contribute zero advantages.
Algorithm~\ref{alg:group-relative-refinement} summarizes the update.

\subsection{Sequential Accept-or-Resample Decoding}

At inference time, \method{} applies a sequential rejection policy at each
committed prefix.
The generator first samples a single candidate step, which is committed when
its acceptance score is at least $\eta$.
Otherwise, the candidate is discarded and a replacement is sampled from the
unchanged prefix, with at most $M$ total candidate attempts at each position.
If every attempt is rejected, the highest-scoring candidate is committed.
An episode-level budget permits at most $B$ rejected candidates; after it is
exhausted, subsequent candidates are committed directly.
Algorithm~\ref{alg:sequential-rejection} specifies this procedure.
Additional candidates are sampled only after a rejection; a candidate accepted
on the first attempt requires no additional candidate generation.
Generation and acceptance scoring share the same backbone, with the
acceptance-gate adapter activated only for scoring
(Appendix~\ref{sec:implementation-details}).

We calibrate $\eta$ before policy refinement to match a prescribed rejection
rate and hold it fixed during refinement and deployment.
This on-policy calibration accounts for the additional samples induced by
rejections and the resulting candidate distribution encountered by the gate.
We consequently report the realized on-policy rejection rate together with its
target.
The two budgets complement one another: $M$ limits concentration of sampling
at one position, while $B$ controls total rejection across the trajectory.

\begin{algorithm}[t]
\caption{Confidence-Qualified First-Passage Labeling}
\label{alg:first-passage-labeling}
\KwIn{Problem $q$; frozen policy $\pi$; sampled trajectory
$p_0,\ldots,p_T$; threshold $\tau$; confidence multiplier $c$;
rollout schedule
$\mathcal K=(K^{(1)},\ldots,K^{(J)})$, where
$K^{(J)}=K_{\max}$.}
\KwOut{Labeled prefixes $\mathcal D$.}

Estimate $\widehat V_0=\widehat V^\pi(q)$\;
\If{$\widehat V_0=0$}{
    \Return{$\varnothing$}\;
}
$\widehat V_{\mathrm{barrier}}
\leftarrow(1-\tau)\widehat V_0$\;
$\mathcal D\leftarrow\varnothing$\;
\For{$t\leftarrow1$ \KwTo $T$}{
    $j\leftarrow1$\;
    Draw $K^{(j)}$ continuations from $p_t$ and estimate $\widehat V_t$\;
    $\mathrm{gap}_t
    \leftarrow\widehat V_{\mathrm{barrier}}-\widehat V_t$\;
    \While{the confidence interval of
    $\mathrm{gap}_t$ contains $0$ and $j<J$}{
        $j\leftarrow j+1$\;
        Draw additional continuations to reach $K^{(j)}$ samples
        and update $\widehat V_t$ and $\mathrm{gap}_t$\;
    }
    \uIf{$\mathrm{gap}_t
    +c\widehat{\mathrm{se}}(\mathrm{gap}_t)<0$}{
        Add $(q,p_t,\textsc{accept})$ to $\mathcal D$\;
    }
    \uElseIf{$\mathrm{gap}_t
    -c\widehat{\mathrm{se}}(\mathrm{gap}_t)>0$}{
        Add $(q,p_t,\textsc{reject})$ to $\mathcal D$ and stop\;
    }
    \Else{
        Mark $p_t$ gray and exclude it from the supervised loss\;
    }
}
\Return{$\mathcal D$}\;
\end{algorithm}

\begin{algorithm}[t]
\caption{Group-Relative Policy Refinement}
\label{alg:group-relative-refinement}
\KwIn{Training problems $\mathcal Q$; frozen generator $\pi_\theta$;
initialized gate policy $\rho_\phi$; group size $G$; threshold $\eta$;
sampling temperature $\mathrm{temperature}$; attempt cap $M$; rejection
budget $B$.}

\ForEach{sampled problem $q\in\mathcal Q$}{
    \For{$e\leftarrow1$ \KwTo $G$}{
        Decode one episode using $\pi_\theta$ and actions sampled from
        $\rho_\phi$, subject to $M$ and $B$\;
        Extract and score the final answer to obtain $R_e\in\{0,1\}$\;
        Store all gate decisions
        $\mathcal T_e=\{(x_{e,t},u_{e,t})\}_t$\;
    }
    $\overline R\leftarrow G^{-1}\sum_{e=1}^{G}R_e$\;
    \For{$e\leftarrow1$ \KwTo $G$}{
        $A_e\leftarrow R_e-\overline R$\;
        Accumulate
        $-A_e\sum_{t\in\mathcal T_e}
        \log\rho_\phi(u_{e,t}\mid x_{e,t})$\;
    }
}
Normalize by the number of stored gate decisions and update $\phi$\;
\end{algorithm}

\begin{algorithm}[t]
\caption{\method{} Sequential Decoding}
\label{alg:sequential-rejection}
\KwIn{Problem $q$; frozen generator $\pi_\theta$; acceptance-gate parameters
$\phi$; threshold $\eta$; maximum candidate attempts $M$ per position;
episode rejection budget $B$; step-token limit
$L_{\mathrm{step}}$; context limit $L_{\mathrm{ctx}}$.}
\KwOut{Final completion.}

$p\leftarrow\varnothing$\;
$n_{\mathrm{reject}}\leftarrow0$\;
\While{$p$ is nonterminal and $L_{\mathrm{ctx}}$ is not exhausted}{
    $\mathcal A\leftarrow\varnothing$\;
    $\mathrm{committed}\leftarrow\mathrm{false}$\;
    \For{$i\leftarrow1$ \KwTo $M$}{
        $L\leftarrow
        \min\{L_{\mathrm{step}},
        L_{\mathrm{ctx}}-\operatorname{tokens}(q,p)\}$\;
        Sample a candidate step $s\sim\pi_\theta(\cdot\mid q,p)$
        with at most $L$ new tokens\;
        \If{$n_{\mathrm{reject}}\ge B$}{
            $p\leftarrow p\oplus s$\;
            $\mathrm{committed}\leftarrow\mathrm{true}$ and break\;
        }
        $\mathrm{score}\leftarrow\operatorname{score}_\phi(q,p,s)$\;
        Add $(s,\mathrm{score})$ to $\mathcal A$\;
        \If{$\mathrm{score}\ge\eta$}{
            $p\leftarrow p\oplus s$\;
            $\mathrm{committed}\leftarrow\mathrm{true}$ and break\;
        }
        $n_{\mathrm{reject}}\leftarrow n_{\mathrm{reject}}+1$\;
    }
    \If{$\mathrm{committed}=\mathrm{false}$}{
        Select the highest-scoring candidate $s^\star$ from $\mathcal A$\;
        $p\leftarrow p\oplus s^\star$\;
    }
}
\Return{$p$}\;
\end{algorithm}

\section{Datasets}
\label{sec:dataset-details}

Our evaluation suite contains 293 contest-mathematics problems from five
test-only benchmarks.
No evaluation problem is used to train the acceptance gate.

\paragraph{AIME 2025.}
We use all 30 problems from the 2025 American Invitational Mathematics
Examination as a single test set, sourced from MathArena
\citep{dekoninck2026beyond}.
The problems cover algebra, combinatorics, geometry, and number theory and
require a three-digit integer final answer.

\paragraph{OlymMATH-Easy/Hard.}
We use all 100 English-language problems from each of the easy and hard
OlymMATH test sets \citep{sun2026challenging}.
Both contain olympiad-style problems with exact symbolic answers spanning
algebra, combinatorics, geometry, and number theory.
Geometry and combinatorics are the largest subject groups in the easy set,
whereas the hard set is balanced with 25 problems per subject.

\paragraph{HMMT February 2025/2026.}
We use all 30 problems from the February 2025 Harvard--MIT Mathematics
Tournament and all 33 from the February 2026 tournament, sourced from
MathArena \citep{dekoninck2026beyond}.
In each year, the first 30 problems are divided evenly among algebra and number
theory, combinatorics, and geometry.
The 2025 answers include integers, fractions, radicals, and other exact
expressions; the 2026 set additionally contains three longer proof-formatted
problems with rule-verifiable final answers.

\paragraph{Training data.}
We construct the acceptance gate's training pool from the 4,428 Omni-MATH
problems \citep{gao2025omni}.
After removing duplicates, we retain rule-verifiable problems with difficulty
levels 4--8, decontaminate them against all 293 evaluation problems, and draw a
stratified sample.
The resulting pool contains 1,100 problems, including those from HMMT February
and November, Putnam, national team selection tests, the IMO Shortlist, and USAMO.
We use 1,004 problems for training and hold out 96 for policy-refinement
development, including early stopping, checkpoint selection, and threshold
calibration.

\section{Implementation Details}
\label{sec:implementation-details}

\subsection{Step and Label Construction}

Unless otherwise specified, Qwen3-14B and
DeepSeek-R1-Distill-Llama-8B use temperature 0.6 and top-$p$ 0.95, while
gpt-oss-20b uses temperature 1.0, top-$p$ 1.0, and
\texttt{reasoning\_effort=medium}.
All generators use a context limit of
$L_{\mathrm{ctx}}=32{,}768$ tokens.

A reasoning step is one generation segment containing at most
$L_{\mathrm{step}}$ newly generated tokens.
Unless otherwise specified, we set $L_{\mathrm{step}}=2048$.
For first-passage labeling (Algorithm~\ref{alg:first-passage-labeling}), we use
$\tau=0.30$, confidence multiplier $c=1$, and rollout schedule
$\mathcal K=(8,12,16)$, with $K_{\max}=16$.
A step may terminate earlier when generation finishes, and its generation
budget is dynamically reduced when fewer than $L_{\mathrm{step}}$ positions
remain in the context window.

During label collection, each evaluated prefix permits at most $M=8$ candidate
attempts. Candidate continuation values are estimated within the remaining
context,
\begin{equation}
    L_{\mathrm{available}}
    =
    L_{\mathrm{ctx}}
    -
    \operatorname{tokens}(q,p_t).
\end{equation}
The resulting confidence-qualified decisions are converted into labels using
the first-passage rule in Appendix~\ref{sec:first-passage-labeling}.
A smaller generation cap is treated as an explicit budgeted condition rather
than as the model's context limit.
For reproducibility, each estimate retains its decoding configuration, seed,
token counts, extracted answer, correctness, finish reason, and exact
continuation or cutoff prefix.

\subsection{Gate Architecture and Training}

\paragraph{Input serialization.}
The acceptance-gate input concatenates the problem, committed prefix,
candidate step, and a fixed marker token.
The dynamically adjusted continuation budget keeps the problem, prefix, and
candidate within $L_{\mathrm{ctx}}$.
The marker is used only for scoring and is excluded from the generation-token
budget.
We use the pretrained reserved token \texttt{<|fim\_pad|>} as the marker and
require that it not occur in any input segment.
The marker is appended at the final input position.

\paragraph{Token-based readout.}
The frozen language-modeling head is evaluated at the marker position.
Following the token-based readout of Math-Shepherd \citep{wang2024math}, the
existing single-token vocabulary entries \texttt{+} and \texttt{-} verbalize
\textsc{accept} and \textsc{reject}, respectively.
If their logits are $\ell^+$ and $\ell^-$, training uses the logit difference
$z=\ell^+-\ell^-$ with binary cross entropy with logits, while inference uses
the equivalent two-token softmax score
$\operatorname{score}
=\exp(\ell^+)/[\exp(\ell^+)+\exp(\ell^-)]$.
The marker embedding, vocabulary embeddings, and language-modeling head remain
frozen; only the LoRA adapter is trained.
Reusing the pretrained language-modeling head provides a stronger inductive
bias under limited gate supervision, for which a newly initialized
classification head readily overfits.

\paragraph{LoRA configuration.}
The only trainable parameters are rank-16 LoRA adapters with scaling factor 32,
zero dropout, and no bias.
They are attached in every transformer layer to
\texttt{q\_proj}, \texttt{o\_proj}, \texttt{gate\_proj},
\texttt{up\_proj}, and \texttt{down\_proj}.
The key and value projections remain frozen.
Generation and acceptance scoring maintain separate incremental key--value
caches over the committed prefix.

Training, development, and test partitions are disjoint at the problem level.
We optimize with AdamW \citep{loshchilov2017decoupled} for three epochs at learning rate
$3\times10^{-4}$, effective batch size eight, and gradient-norm clipping at
1.0.
Class-dependent weighting addresses the imbalance in first-passage labels.
For same-prefix accept/reject pairs, the pointwise losses are augmented with
$\lambda_{\mathrm{pair}}\operatorname{softplus}[-(z_a-z_r)]$, with
$\lambda_{\mathrm{pair}}=0.5$ and
$\operatorname{softplus}(x)=\log(1+\exp x)$.
This smooth ranking loss encourages a higher logit for the accepted candidate.
Checkpoint selection uses within-prefix discrimination and accepted-value
lift.

\subsection{Policy-Refinement Configuration}

Policy refinement uses $G=8$ episodes per problem and
$\mathrm{temperature}=1.0$.
We set the target rejection rate to 17.8\%, based on the rate measured under
oracle-gated decoding.
Before refinement, we calibrate $\eta$ to the corresponding empirical quantile
of the initial adapter's on-policy acceptance scores.
The resulting threshold remains fixed throughout policy refinement and
deployment; the realized rejection rate is reported separately in
Table~\ref{tab:deployment-behavior}.

Each iteration draws 16 fresh training problems, sampling without replacement
until the pool is exhausted and then reshuffling it.
The resulting groups are combined into an effective batch of 256 gate
decisions, and one AdamW update is applied at learning rate
$5\times10^{-5}$, zero weight decay, and gradient-norm clipping at 1.0.
Generation uses temperature 0.6 and top-$p$ 0.95 for Qwen3-14B and
DeepSeek-R1-Distill-Llama-8B, and temperature 1.0, top-$p$ 1.0, and
\texttt{reasoning\_effort=medium} for gpt-oss-20b.
All generators use 2048-token reasoning steps and a context limit of 32,768
tokens.
Each position permits $M=8$ total candidate attempts, and each episode permits
$B=12$ rejected candidates.
When the physical context is exhausted, we extract and score any available
final answer; an episode with no extractable answer receives reward zero.

\begin{figure}[t]
\centering
\includegraphics[width=0.9\linewidth]{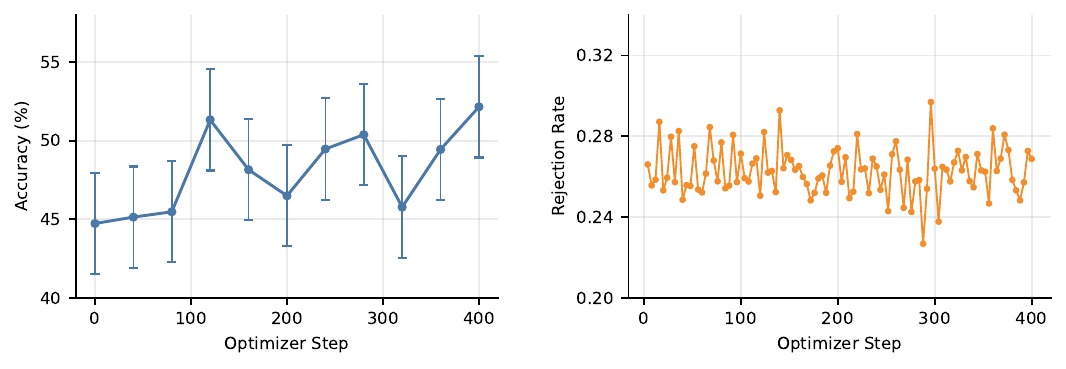}
\caption{Held-out development accuracy (left) and realized on-policy rejection
rate (right) over optimizer steps during policy refinement with Qwen3-14B.}
\label{fig:rl-training-dynamics}
\end{figure}

Training is bounded by a preset compute budget and development early stopping.
Development evaluation samples four episodes for each of 96 held-out
Omni-MATH problems.
An evaluation more than two standard errors below the historical best is
marked as degradation.
Training stops after two consecutive degraded evaluations, and the checkpoint
with the highest development accuracy is selected.
Figure~\ref{fig:rl-training-dynamics} summarizes the optimization trajectory:
held-out accuracy follows a noisy upward trend, while the realized on-policy
rejection rate remains broadly stable under the fixed $\eta$.
Training uses a single H200 node, with a total wall-clock training time of
approximately 200 hours per model.

\subsection{Inference Runtime}

Candidate generation runs with the LoRA adapter disabled through
the adapter context, preserving the frozen generator exactly.
Candidate evaluation enables the adapter and reads only the final marker
position.
Because the marker is guaranteed to be last, we request one output position
instead of materializing logits of shape
sequence-length by vocabulary-size.
Generation and acceptance scoring each maintain an incremental key--value
cache over committed steps.
Scoring a new candidate processes only its new tokens and the appended marker
rather than recomputing the full trajectory.
Rejected candidates use temporary cache branches that are discarded after
scoring.
Table~\ref{tab:deployment-behavior} shows that the learned gate rejects
selectively: the per-position attempt cap is reached on only 0.4\% of
committed steps, and the episode budget is exhausted on 3.0\% of problems.
As shown in Table~\ref{tab:wall-clock-runtime}, \methodabbr{} adds 0.17 hours
over single-pass decoding while remaining substantially faster than
full-solution scaling.

\begin{table*}[t]
\centering
\begin{minipage}[t]{0.53\textwidth}
\centering
\scriptsize
\setlength{\tabcolsep}{5pt}
\caption{\methodabbr{} deployment behavior on Qwen3-14B, averaged over three
evaluation runs.}
\label{tab:deployment-behavior}
\begin{tabular}{lc}
\toprule
Metric & Mean \\
\midrule
Candidate rejection rate & 21.0\% \\
Candidate attempts per committed step & 1.3 \\
Committed steps per problem & 8.82 \\
Gate evaluations per problem & 10.7 \\
Steps reaching the $M$-attempt cap & 0.4\% \\
Problems exhausting the $B$-rejection budget & 3.0\% \\
\bottomrule
\end{tabular}
\end{minipage}
\hfill
\begin{minipage}[t]{0.41\textwidth}
\centering
\scriptsize
\setlength{\tabcolsep}{8pt}
\caption{Qwen3-14B wall-clock time per evaluation run, with generation on six
H200 GPUs.}
\label{tab:wall-clock-runtime}
\begin{tabular}{lc}
\toprule
Method & Wall-clock time (h) \\
\midrule
Single pass & 1.09 \\
\methodabbr{} & 1.26 \\
ESC ($w=2$) & 4.85 \\
Majority@8 & 8.81 \\
USC & 8.95 \\
\bottomrule
\end{tabular}
\end{minipage}
\end{table*}

\subsection{Evaluation Metrics}
\label{sec:evaluation-metrics}

\paragraph{Gate metrics.}
Global AUROC is insufficient because it rewards problem-difficulty prediction.
Our primary offline metrics are within-prefix pairwise AUROC, rejection
precision at a fixed rejection rate, and the accepted-value lift
\begin{equation}
    \Delta V_{\mathrm{accept}}
    =
    \mathbb E[V^\pi(q,p\oplus s)\mid s\ \text{accepted}]
    -
    \mathbb E_{s\sim\pi(\cdot\mid q,p)}[V^\pi(q,p\oplus s)].
\end{equation}
We evaluate four baselines.
Length+depth is a fitted two-feature probe using candidate token length and
committed-prefix depth.
Log-prob uses the frozen generator's mean per-token log-probability from a
teacher-forced pass over the fully rendered prompt and requires no fitting.
Frozen MLP is a one-hidden-layer probe trained on standardized, PCA-reduced
final-token hidden states from a Transformer layer fixed before evaluation,
with the prompt included.
The Skywork verifier directly applies Skywork-Reward-V2-Qwen3-8B to the complete problem,
prefix, and candidate sequence without task-specific training.
All fitted probes are trained only on problem-grouped training folds.

\paragraph{End-to-end metrics.}
We report final-answer accuracy, generated-token cost relative to single-pass
generation, candidate rejection rate, candidate attempts per committed step,
$M$-cap fallback rate, $B$-budget exhaustion rate, and end-to-end wall-clock
time.
All online policies are evaluated in paired runs with deterministic
problem/sample seeds.
Threshold $\eta$, attempt cap $M$, episode rejection budget $B$, and checkpoint
are selected on development data only.

\begin{table*}[t]
\centering
\begin{minipage}[t]{0.48\textwidth}
\centering
\scriptsize
\setlength{\tabcolsep}{3.5pt}
\caption{Measured \methodabbr{} sensitivity to the target rejection rate on
the test suite with Qwen3-14B and $L_{\mathrm{step}}=2048$. Results are
averaged over three evaluation runs.}
\label{tab:rejection-rate-sensitivity}
\begin{tabular}{ccc}
\toprule
Target rej.\ (\%) & Tok.$\times$ & Accuracy (\%) \\
\midrule
7.5  & 1.07 & 60.3 \\
10.0 & 1.10 & 61.0 \\
12.5 & 1.14 & 62.1 \\
15.0 & 1.17 & 63.2 \\
\rowcolor{gray!15}
17.8 & 1.21 & 64.6 \\
22.0 & 1.27 & 64.6 \\
30.0 & 1.37 & 64.7 \\
\bottomrule
\end{tabular}
\end{minipage}
\hfill
\begin{minipage}[t]{0.48\textwidth}
\centering
\scriptsize
\setlength{\tabcolsep}{3.5pt}
\caption{Measured \methodabbr{} sensitivity to $L_{\mathrm{step}}$ on the test
suite with Qwen3-14B at a fixed target rejection rate of 17.8\%. Results are
averaged over three evaluation runs.}
\label{tab:step-sensitivity}
\begin{tabular}{cccc}
\toprule
$L_{\mathrm{step}}$ & Avg.\ steps & Tok.$\times$ & Accuracy (\%) \\
\midrule
256  & 63.3 & 1.27 & 54.7 \\
512  & 31.7 & 1.27 & 54.4 \\
1024 & 15.8 & 1.26 & 64.6 \\
\rowcolor{gray!15}
2048 & 8.82 & 1.21 & 64.6 \\
4096 & 4.1 & 1.23 & 64.1 \\
8192 & 2.3 & 1.21 & 62.2 \\
\bottomrule
\end{tabular}
\end{minipage}
\end{table*}

\section{Additional Results}
\label{sec:additional-results}

\subsection{Power Analysis for Local Value Differences}
\label{sec:diagnostic-cases}

For two Bernoulli proportions near success probability $\mu$, the approximate
number of rollouts per endpoint required to resolve an effect $\delta$ with
80\% power is
$n\approx2(1.96+0.84)^2\mu(1-\mu)/\delta^2$.
Using the magnitude of the mean adjacent change rounded to $\delta=0.05$,
the worst-case $\mu=0.5$ calculation gives approximately 1,600 rollouts per
endpoint.

\subsection{Sensitivity Analysis}
\label{sec:sensitivity-analysis}

All sensitivity results in this subsection are measured with Qwen3-14B.

\paragraph{Target rejection rate.}
Table~\ref{tab:rejection-rate-sensitivity} varies the target rejection rate
and calibrates $\eta$ for each operating point.
Accuracy improves as the gate is allowed to reject more candidates, then
plateaus across target rates from 17.8\% to 30.0\% while token cost continues
to increase.
The default target of 17.8\% lies near the start of this plateau.

\paragraph{Step size.}
\label{sec:step-sensitivity}

Table~\ref{tab:step-sensitivity} varies $L_{\mathrm{step}}$ with the target
rejection rate fixed at 17.8\%.
Very short steps create many decision points and increase scoring and
resampling cost.
Very long steps delay intervention and yield lower accuracy without reducing
generated-token cost below the 2048-token setting.
Step sizes of 1,024--4,096 tokens achieve similar accuracy, with the default
$L_{\mathrm{step}}=2048$ using the fewest generated tokens in this range.

\subsection{Examples}
\label{sec:ssr-examples}

We present two measured \methodabbr{} trajectories that yield correct answers
where the corresponding single-pass runs fail.
On AIME 2025 Problem 10, Qwen3-14B
rejects two candidates, one introducing an unstated structural constraint and
the other leaving the counting argument unresolved, and returns the correct
answer, $81$
(Table~\ref{tab:ssr-example-qwen}). On HMMT February 2026 Problem 19,
gpt-oss-20b rejects a faulty construction, replaces it with the nesting property
$A_d\subseteq A_n$ used by the remainder of the solution, and returns the
correct answer, $499$
(Table~\ref{tab:ssr-example-gptoss}).

\begin{table*}[t]
\centering
\small
\caption{Abridged SSR trajectory for AIME 2025 Problem 10 with Qwen3-14B.
Two candidates are rejected; omitted steps are accepted on their first attempt.}
\label{tab:ssr-example-qwen}
\setlength{\tabcolsep}{4pt}
\begin{tabular}{@{}clp{0.58\textwidth}@{}}
\toprule
Step & Decision & Abridged candidate reasoning \\
\midrule
$t=1$ & Accept &
Identifies the row and $3\times3$-block constraints and begins reducing the
problem to a combinatorial count. \\
\midrule
$t=2$, cand.\ 1 & Reject &
Incorrectly infers that entries must also be distinct within each column of a
block and repeatedly recasts the block as a Latin-square-like object. \\
$t=2$, cand.\ 2 & Accept &
Retains only the stated row and block constraints and represents each block by
an ordered partition of $\{1,\ldots,9\}$ into three row subsets. \\
\midrule
$t=3$--$5$ & Accept &
\(\cdots\) Reduces the remaining task to counting the assignments for the
second and third blocks after fixing the first block. \\
\midrule
$t=6$, cand.\ 1 & Reject &
Formulates the assignment as a constrained bipartite matching problem but does
not obtain the required count. \\
$t=6$, cand.\ 2 & Accept &
Derives the row-assignment count
$\sum_{c=0}^{3}\binom{3}{c}^{3}=56$ and obtains
$9!\cdot56\cdot(3!)^6$. \\
\midrule
$t=7$--$10$ & Accept &
\(\cdots\) Verifies
$9!\cdot56\cdot(3!)^6=2^{16}3^{10}5\,7^2$ and returns
$2(16)+3(10)+5(1)+7(2)=\boxed{81}$. \\
\bottomrule
\end{tabular}
\end{table*}

\begin{table*}[t]
\centering
\small
\caption{Abridged SSR trajectory for HMMT February 2026 Problem 19 with
gpt-oss-20b. One candidate is rejected; omitted steps are accepted on their
first attempt.}
\label{tab:ssr-example-gptoss}
\setlength{\tabcolsep}{4pt}
\begin{tabular}{@{}clp{0.58\textwidth}@{}}
\toprule
Step & Decision & Abridged candidate reasoning \\
\midrule
$t=1$ & Accept &
Reads $|A_i|=i$ from the case $i=j$ and restates the objective as minimizing the
eight maxima $\max A_d$ over the divisors of $250$. \\
\midrule
$t=2$ & Accept &
Observes that the elements are arbitrary positive integers chosen by the
construction, so only which labels are assigned to which sets matters. \\
\midrule
$t=3$, cand.\ 1 & Reject &
Assigns one element $e_t$ to every $A_k$ with $t\mid k$, computes that this
yields $|A_i\cap A_j|\ge d(\gcd(i,j))$ rather than $\gcd(i,j)$, and on finding
the count too small begins a prime-by-prime case analysis instead of revising the
construction. \\
$t=3$, cand.\ 2 & Accept &
Derives the structural lemma: for $d\mid n$, $|A_n\cap A_d|=d=|A_d|$ forces
$A_d\subseteq A_n$, so the family is nested along divisibility. Assigns each
element its minimal index $m$ and concludes $e\in A_n$ iff $m\mid n$. \\
\midrule
$t=4$--$7$ & Accept &
\(\cdots\) Writes $|A_n|=\sum_{m\mid n}f(m)=n$ for the count $f(m)$ of elements
of minimal index $m$, giving
$f=1,1,4,4,20,20,100,100$ on the divisors of $250$, and reduces the objective to
choosing each group's largest label $g(m)$, since
$\max A_d=\max_{m\mid d}g(m)$. \\
\midrule
$t=8$--$12$ & Accept &
\(\cdots\) Evaluates orderings of the eight groups. Labeling them in ascending
order of size gives
$g=1,2,6,10,30,50,150,250$ and the sum $499$, while moving index $5$ earlier
raises the total to $502$. \\
\midrule
$t=13$ & Accept &
Terminal step. Rules out the smaller candidate value $468$, which would require
$g(d)=d$ and violates the monotonicity $g(m)<g(n)$ for $m\mid n$, and returns
$\boxed{499}$. \\
\bottomrule
\end{tabular}
\end{table*}

\end{document}